# Self-Reported Confidence of Large Language Models in Gastroenterology: Analysis of Commercial, Open-Source, and Quantized Models

**Authors:** Nariman Naderi[1†], Seyed Amir Ahmad Safavi-Naini[1,2†], Thomas Savage[3], Zahra Atf[4], Peter Lewis[4], Girish Nadkarni[1,2], Ali Soroush[1,2,5*]

**Affiliations:**

1- Division of Data-Driven and Digital Medicine (D3M), Department of Medicine, Icahn School of Medicine at Mount Sinai, New York, NY, USA.

2- The Charles Bronfman Institute of Personalized Medicine, Icahn School of Medicine at Mount Sinai, New York, New York, USA.

3- Division of Hospital Medicine, University of Pennsylvania, Pittsburg, USA.

4- Faculty of Business and Information Technology, Ontario Tech University, Oshawa, Canada.

5- Henry D. Janowitz Division of Gastroenterology, Department of Medicine, Icahn School of Medicine at Mount Sinai, New York, NY, USA.

† Nariman Naderi and Seyed Amir Ahmad Safavi-Naini contributed equally to this work.

**Corresponding author:** Ali Soroush, M.D., M.S. (ali.soroush@mountsinai.org)

# Abstract

This study evaluated self-reported response certainty across several large language models (GPT, Claude, Llama, Phi, Mistral, Gemini, Gemma, and Qwen) using 300 gastroenterology board-style questions. The highest-performing models (GPT-o1 preview, GPT-4o, and Claude-3.5-Sonnet) achieved Brier scores of 0.15-0.2 and AUROC of 0.6. Although newer models demonstrated improved performance, all exhibited a consistent tendency towards overconfidence. Uncertainty estimation presents a significant challenge to the safe use of LLMs in healthcare.

Large Language Models (LLMs) have the potential to transform healthcare and are increasingly being used for tasks such as clinical decision support and patient data interpretation [1]. However, they are prone to hallucinations, presenting incorrect information with convincing terminology and overconfidence [2,3]. Hallucinations in the medical applications of LLMs are particularly concerning, as these models can present incorrect information with sophisticated terminology and apparent confidence, which may mislead individuals without specialized medical expertise[4]. This limitation poses significant risks in clinical settings, where output accuracy is critical[5]. Consequently, quantifying model uncertainty has become essential for safe clinical implementation, as confidence and uncertainty metrics enhance trust, explainability, and overall usability of LLMs in high-stakes domains such as medicine [6].

Five distinct approaches exist for assessing uncertainty without specialized datasets. Token-level probability analysis evaluates the model confidence in generating each token or text segment[7–9]. However, this approach is computationally intensive for long-form texts. Semantic difference assessments compare differences in meaning across multiple LLM outputs[6,10–13]. However, this method requires multiple generations for each response, increasing both the cost and latency. Machine learning (ML)-based approaches utilize features from the prompt and the output of LLM to train an ML model for generating future prompts and confidence estimation, [14,15]. These approaches require vast training datasets and often lack generalizability to new contexts. Surrogate model approaches combine the verbalized confidence of a robust closed-source LLM with the token logits of a weaker model for uncertainty quantification. Although this method has shown promise, it may produce unexpected results when the outputs of the two models diverge significantly. Finally, confidence elicitation uses the natural language capabilities of LLMs to self-report response certainty[16–19]. This approach leverages the generalized knowledge embedded within LLMs to provide a more practical solution. However, our previous research revealed that LLMs, including GPT-4 and Llama-3, fail to accurately represent their confidence, raising concerns about the use of this approach[20].

LLM architectures and training datasets vary significantly across model families, potentially unpredictably impacting confidence elicitation performance. However, the confidence elicitation performance of LLMs has not been fully characterized, particularly for open-source models that offer enhanced flexibility and data security for medical applications. In addition to architectural and dataset variations, the breadth of knowledge embedded in LLMs across various topics can be influenced by differences in training dataset sizes. This is especially true for specialized fields, such as board-level questions in gastroenterology, where these discrepancies may introduce unknown effects on confidence elicitation processes. This warrants further investigation, particularly when contrasted with the already researched and more general fields of confidence elicitation. This study systematically evaluated the self-reported confidence elicitation of different commercial and open-source LLMs across local, web, and API-based environments.

To assess LLM confidence elicitation, we used the 2022 American College of Gastroenterology self-assessment examination, which contains 300 board exam-style multiple-choice questions and answers written by expert gastroenterologists. Access to these examination questions is typically restricted behind paywalls, making them suitable test cases for assessing LLM performance owing to their likely exclusion from training datasets. Furthermore, gastroenterology is a subspecialty of medicine and demonstrates confidence elicitation performance when evaluating the boundaries of model knowledge.

We utilized our previous LLM answer generation design[21], incorporating an initial step to determine the optimized model parameters (prompt-engineered prompt, temperature, and max-token). Response accuracy was calculated based on the LLM-selected option and its alignment with the correct option. A semi-automated approach was employed for LLM answer extraction and evaluation, with human-in-the-loop verification achieving a 99% accuracy.

Confidence was reported as a value between 0 (least confidence) and 10 (most confidence). For an initial analysis, the mean confidence and mean accuracy of all models were calculated to provide a global view of LLMs' potential for confidence elicitation. Subsequently, the trends in confidence scores were analyzed with respect to question difficulty (percentage of test takers answering a question correctly) and question length. The rationale behind these analyses is that a well-performing model should exhibit lower confidence when its accuracy is low and should demonstrate decreasing confidence scores as the question difficulty increases. Model discrimination and the calibration of model certainty were assessed using AUROC and Brier score, respectively. **Figure 1** illustrates the study design, including answer generation, confidence score extraction, and analysis methodology.

Forty-eight LLMs comprising 6 quantized and 42 base models across seven different model families were evaluated. These included Qwen (2, 2.5), Llama (3, 3.1, 3.2, 3.3), Phi (3, 3.5), Gemini (1.0 (normal, Pro)), Gemma (2), Claude (3, 3.5), and GPT (3.5, 4, 4o, 4omini, o1, o1mini). A total of 13,362 options and 12,307 self-reported confidence scores were successfully extracted from unstructured responses. The discrepancy between the number of generated responses and the accompanying confidence scores can be attributed to two primary factors. First, certain LLMs failed to adhere to prompt instructions, resulting in the omission of confidence scores. Second, the limitation on the maximum number of output tokens (512 tokens in addition to the input tokens) led to the premature truncation of responses by some LLMs. This issue was particularly evident in reasoning models, likely because they allocated a substantial number of tokens to reasoning processes before generating a final answer, leaving insufficient tokens to include confidence scores.

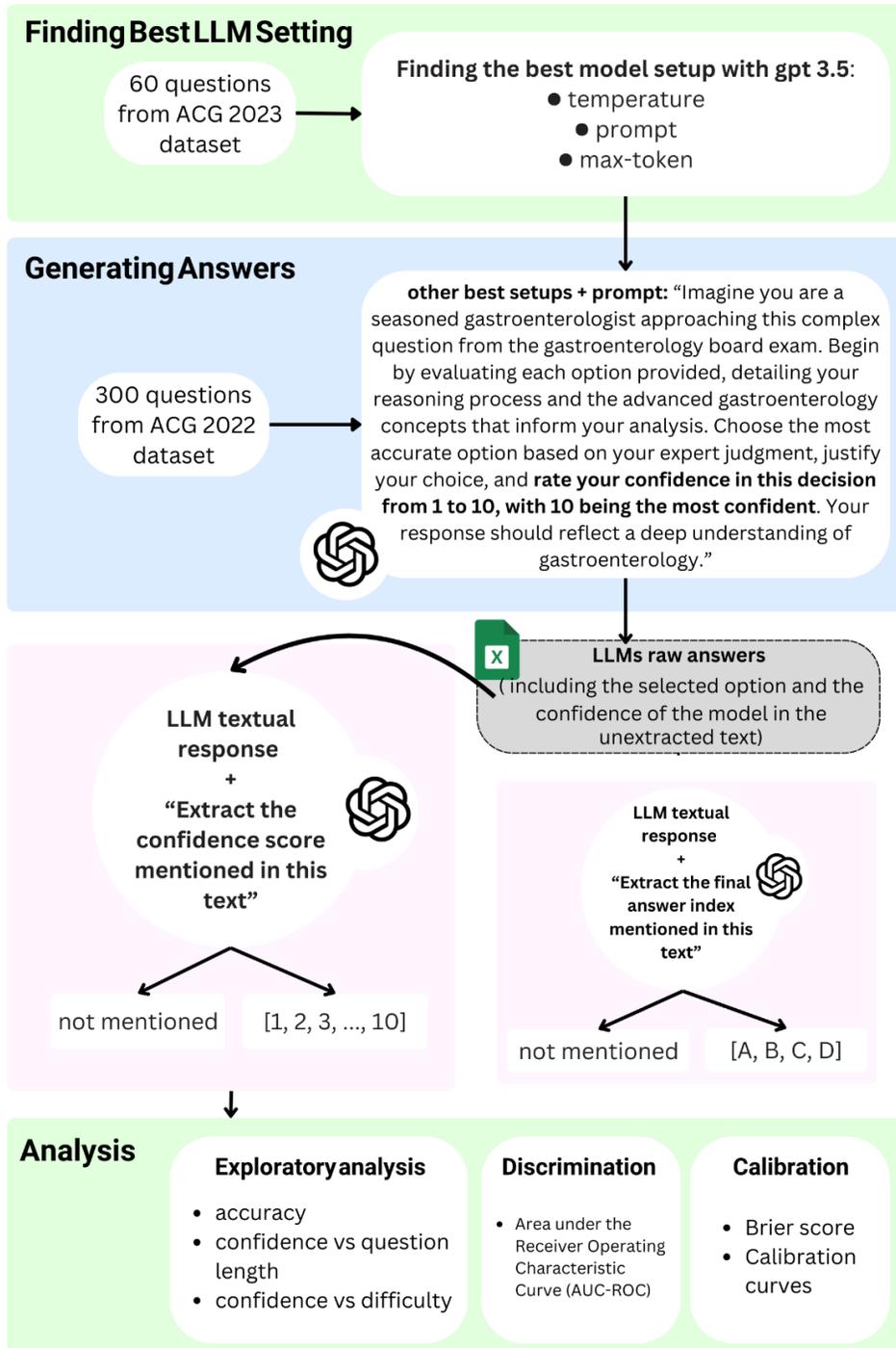

**Figure 1. Summary illustration of pipeline for confidence score extraction from raw textual responses.**

The mean confidence scores ranged from 7.99 (95% CI: 7.89-8.09) to 9.58 (95% CI: 9.45) for Claude-3-Opus and Mistral-7b, respectively. The GPT-o1 preview had the best response accuracy (81.5%) and Llama3-8b-Q8 had the worst accuracy (30.3%), consistent with prior findings[21]. **Figure 2** provides a comprehensive overview of LLMs' confidence elicitation capabilities, illustrating the relationship between the overall average confidence scores and overall accuracy. This analysis revealed that the average confidence of the models consistently exceeded their average accuracy, indicating overconfidence. Models positioned closer to the midline are predominantly more recent and have larger parameter sizes.

**Figure 3** shows the distribution of confidence scores relative to average accuracy, stratified by correct and incorrect questions. This highlights the discrepancies between confidence and accuracy, showing that confidence scores for incorrect responses are often similar to those for correct responses.

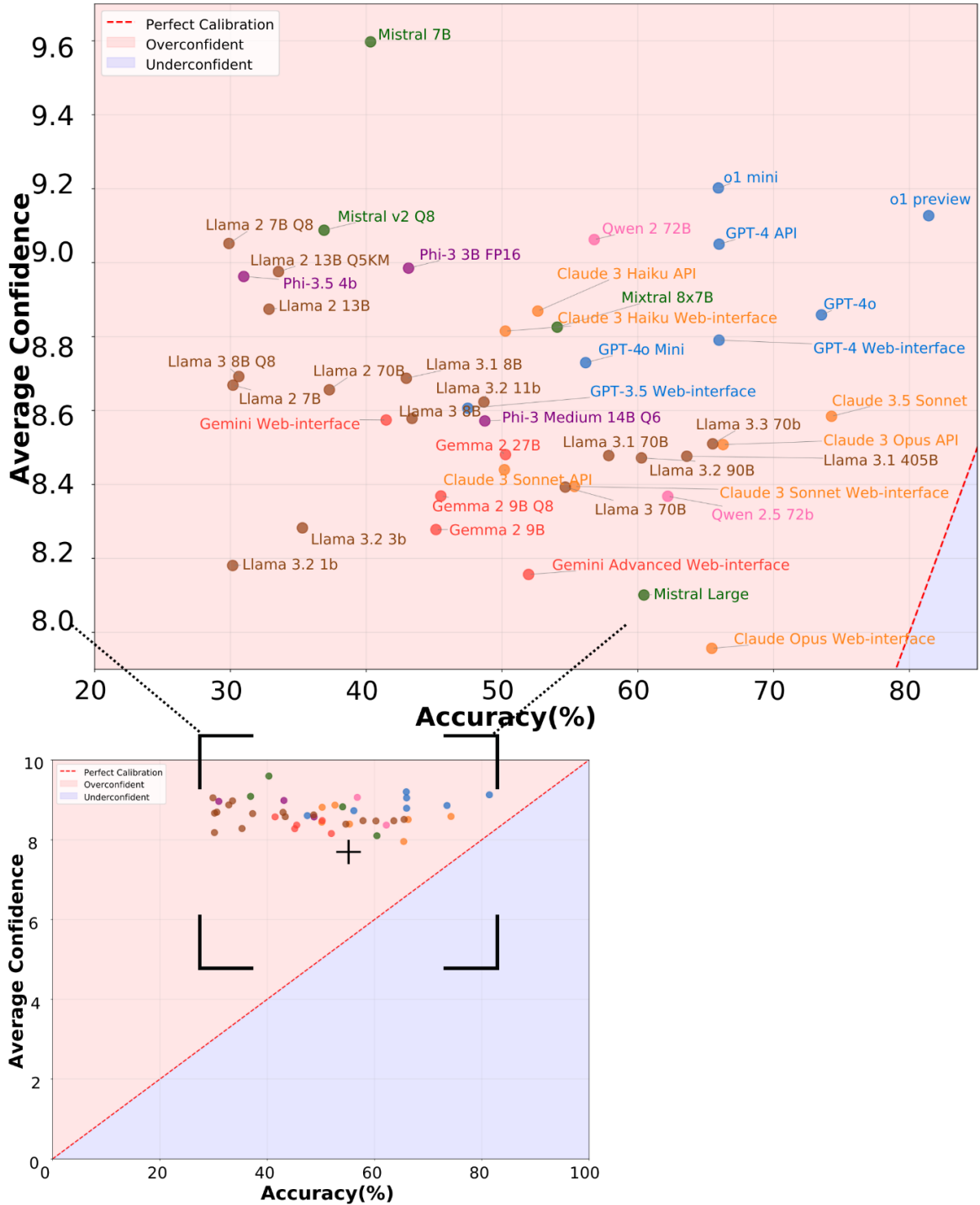

**Figure 2. Average accuracy versus average confidence scores for LLMs with more than 150 valid samples.** The dashed red line indicates perfect calibration, that is, the alignment of the average accuracy and average confidence score. Models above this line are overconfident, whereas those below are under-confident. A subset of the data was magnified for clarity purposes.

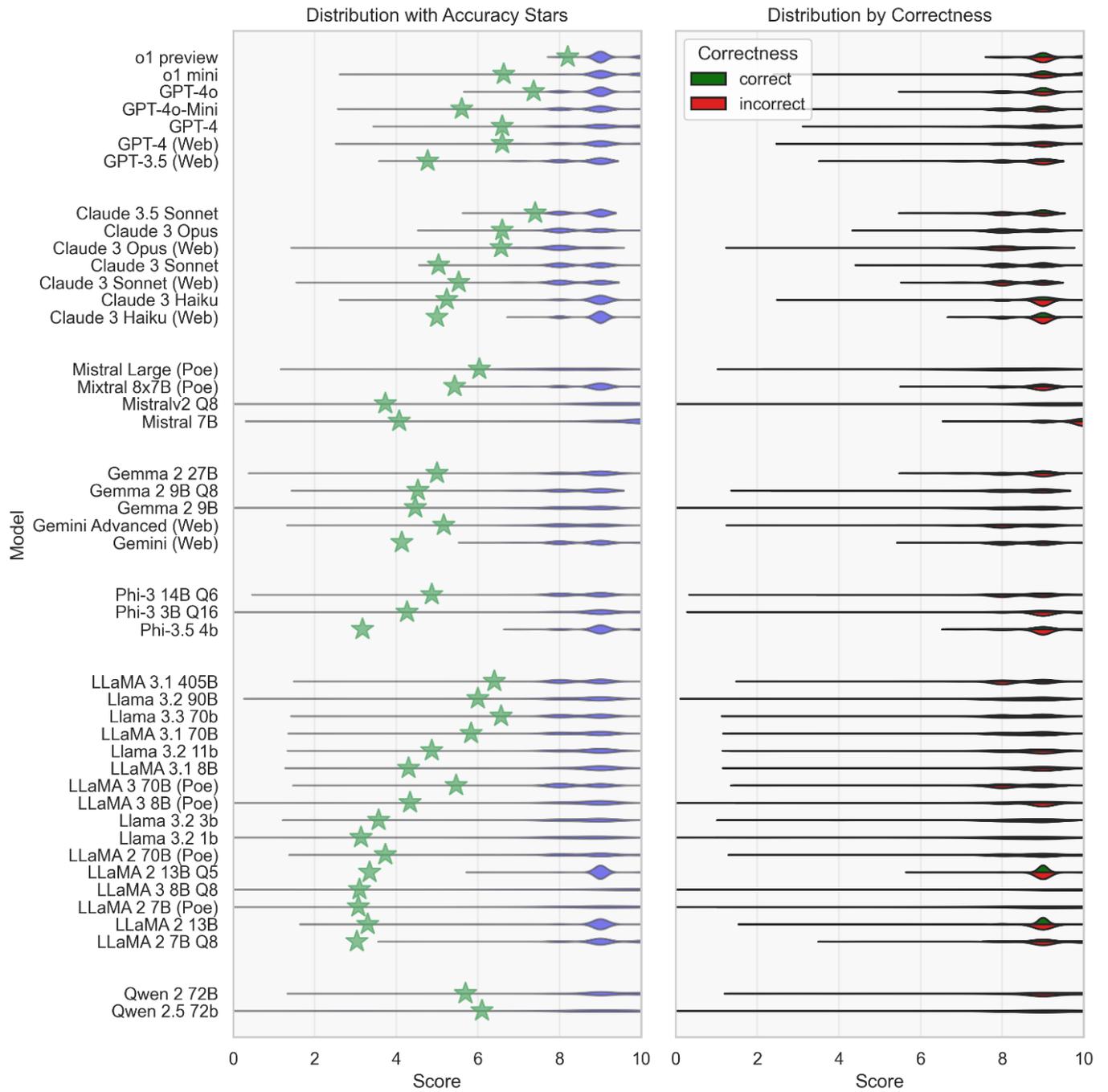

**Figure 3. Left panel:** Overall distribution of self-reported confidence scores and mean response accuracy (stars) for each model. **Right panel:** Distribution of self-reported confidence scores for each model stratified by response accuracy.

The performance metrics for discrimination (AUROC) and calibration (Brier score and ECE) are listed in **Table 1**. GPT-o1 mini achieved the highest AUROC (0.626), followed by GPT-4 (0.605), GPT-4o (0.604), Mistral Large (0.602), and Claude 3.5 Sonnet and Llama-3.2-90b (0.600 for both). Notably, the AUROC of all models was lower than the threshold of 0.7 **(Supplementary Figure S1)**. Only five models demonstrated superior-to-random Brier scores: GPT-o1-preview (0.157), Claude-3.5-Sonnet (0.202), GPT-4o (0.206), Claude-3-Opus-API (0.226), and Claude-3-Opus-Web (0.244) **(Supplementary Figure S2).** Regarding ECE, the GPT-o1-preview exhibited the lowest value (0.100), followed by Claude 3.5 Sonnet (0.122), GPT-4o (0.148), Claude 3 Opus API (0.150), and Claude Opus Web-interface (0.154), indicating better calibration performance **(Supplementary Figure S3).**

| Model family | Model name and parameter (quantization) | Date accessed | Calibration | | Discrimination | Accuracy | Self-reported confidence score |
|---|---|---|---|---|---|---|---|
| | | | Brier score | ECE | AUROC | Percent | Mean (95CI) |
| Llama | | | | | | | |
| | Llama-3.3-70b | December 2024 | 0.260 | 0.199 | 0.563 | 65.66 | 8.46 (8.36-8.56) |
| | Llama 3.1 405B | August 2024 | 0.273 | 0.211 | 0.592 | 64 | 8.47 (8.38-8.57) |
| | Llama3.2-90B | December 2024 | 0.302 | 0.269 | 0.600 | 60.00 | 8.49 (8.34-8.62) |
| | Llama 3.1 70B | August 2024 | 0.313 | 0.283 | 0.538 | 58.19 | 8.51 (8.39-8.62) |
| | Llama 3 70B | May 2024 | 0.334 | 0.301 | 0.572 | 54.66 | 8.38 (8.28-8.48) |
| | Llama 3 8B | May 2024 | 0.422 | 0.450 | 0.478 | 43.33 | 8.54 (8.41-8.68) |
| | Llama-3.2-11b | December 2024 | 0.400 | 0.390 | 0.519 | 48.65 | 8.59 (8.46-8.69) |
| | Llama 3.1 8B | August 2024 | 0.433 | 0.441 | 0.512 | 43.14 | 8.67 (8.54-8.80) |
| | Llama-3.2-3b | December 2024 | 0.465 | 0.487 | 0.534 | 35.66 | 8.32 (8.18-8.45) |
| | Llama 2 70B | April 2024 | 0.481 | 0.493 | 0.529 | 37.71 | 8.70 (8.58-8.81) |
| | Llama-3.2-1b | December 2024 | 0.500 | 0.511 | 0.455 | 30.61 | 8.13 (7.96-8.31) |
| | Llama 2 13B (Q5) | April 2024 | 0.525 | 0.546 | 0.5 | 35.16 | 8.98 (8.92-9.04) |
| | Llama 3 8B (Q8) | April 2024 | 0.527 | 0.613 | 0.472 | 30.35 | 8.65 (8.28-9.02) |
| | Llama 2 7B | April 2024 | 0.528 | 0.587 | 0.47 | 30.87 | 8.66 (8.47-8.84) |
| | Llama 2 13B | April 2024 | 0.531 | 0.558 | 0.52 | 33.11 | 8.89 (8.82-8.95) |
| | Llama 2 7B (Q8) | April 2024 | 0.559 | 0.582 | 0.458 | 32.45 | 9.07 (8.98-9.15) |
| Qwen | | | | | | | |
| | Qwen-2.5-72b | September 2024 | 0.326 | 0.304 | 0.549 | 61.48 | 8.39(8.15-8.63) |
| | Qwen-2-72B | September 2024 | 0.364 | 0.360 | 0.583 | 57.00 | 9.10(8.98-9.20) |
| Phi | | | | | | | |
| | Phi-3 Medium 14B (Q6) | April 2024 | 0.389 | 0.377 | 0.588 | 48.66 | 8.57 (8.48-8.67) |
| | Phi-3 3B FP16 | April 2024 | 0.458 | 0.464 | 0.486 | 43.79 | 8.96 (8.84-9.07) |
| | Phi-3.5-4b | December 2024 | 0.558 | 0.578 | 0.465 | 31.86 | 8.96 (8.90-9.02) |
| Google | | | | | | | |
| | Gemini Advanced Web | March-April 2024 | 0.297 | 0.247 | 0.561 | 58.49 | 8.20 (8.07-8.33) |
| | Gemma 2 27B | July 2024 | 0.374 | 0.352 | 0.557 | 50 | 8.52 (8.41-8.63) |
| | Gemma 2 9B (Q8) | July 2024 | 0.397 | 0.392 | 0.543 | 45.33 | 8.40 (8.30-8.50) |

|  | Model | Date | | | | | |
|---|---|---|---|---|---|---|---|
|  | Gemma 2 9B | July 2024 | 0.398 | 0.390 | 0.592 | 44.59 | 8.33 (8.20-8.45) |
|  | Gemini Web | March 2024 | 0.421 | 0.420 | 0.563 | 44.44 | 8.61 (8.53-8.70) |
| Mistral | | | | | | | |
|  | Mistral Large | April 2024 | 0.282 | 0.224 | 0.602 | 60.53 | 8.13 (7.98-8.28) |
|  | Mixtral 8x7B | April 2024 | 0.359 | 0.336 | 0.547 | 54.33 | 8.79 (8.72-8.87) |
|  | Mistral v2 Q8 | April 2024 | 0.506 | 0.527 | 0.554 | 39.06 | 9.11 (8.90-9.32) |
|  | Mistral 7B | April 2024 | 0.547 | 0.551 | 0.519 | 40.66 | |
| Claude | | | | | | | |
|  | Claude 3.5 Sonnet | July 2024 | 0.207 | 0.122 | 0.6 | 74 | 8.60 (8.54-8.67) |
|  | Claude 3 Opus | March-April 2024 | 0.229 | 0.150 | 0.575 | 70.35 | 8.54 (8.44-8.63) |
|  | Claude 3 Opus Web | March-April 2024 | 0.246 | 0.154 | 0.578 | 65.66 | 7.99 (7.89-8.09) |
|  | Claude 3 Sonnet Web | March-April 2024 | 0.326 | 0.284 | 0.551 | 55.33 | 8.37 (8.29-8.45) |
|  | Claude 3 Sonnet | March-April 2024 | 0.361 | 0.336 | 0.559 | 51.17 | 8.48 (8.39-8.58) |
|  | Claude 3 Haiku | March-April 2024 | 0.373 | 0.357 | 0.522 | 53.76 | 8.88 (8.80-8.96) |
|  | Claude 3 Haiku Web | March-April 2024 | 0.398 | 0.385 | 0.523 | 50 | 8.85 (8.80-8.90) |
| GPT | | | | | | | |
|  | GPT-o1 preview | September 2024 | 0.157 | 0.100 | 0.576 | 81.57 | 9.15 (9.10-9.20) |
|  | GPT-4o | May 2024 | 0.208 | 0.148 | 0.604 | 74 | 8.86 (8.80-8.92) |
|  | GPT-4 Web | March 2024 | 0.267 | 0.221 | 0.588 | 66.22 | 8.79 (8.70-8.87) |
|  | GPT-4 | March 2024 | 0.278 | 0.237 | 0.605 | 66.53 | 9.02 (8.92-9.13) |
|  | GPT-o1 Mini | September 2024 | 0.278 | 0.257 | 0.626 | 66.33 | 9.20 (9.12-9.27) |
|  | GPT-4o Mini | July 2024 | 0.342 | 0.309 | 0.572 | 56.61 | 8.75 (8.67-8.83) |
|  | GPT-3.5 Web | March 2024 | 0.394 | 0.375 | 0.546 | 47.66 | 8.56 (8.48-8.63) |

**Table 1. LLM accuracy, discrimination, calibration, and confidence scores were sorted from best calibration (lowest Brier score) to worst for each model family.**

**Figure 4** presents the calibration curves for the top six and bottom three models with respect to Brier scores. As shown, all models tended to overestimate their confidence relative to their actual knowledge, as reflected in their average accuracies. The calibration curves for all models are presented in **Supplementary Figure S5.**

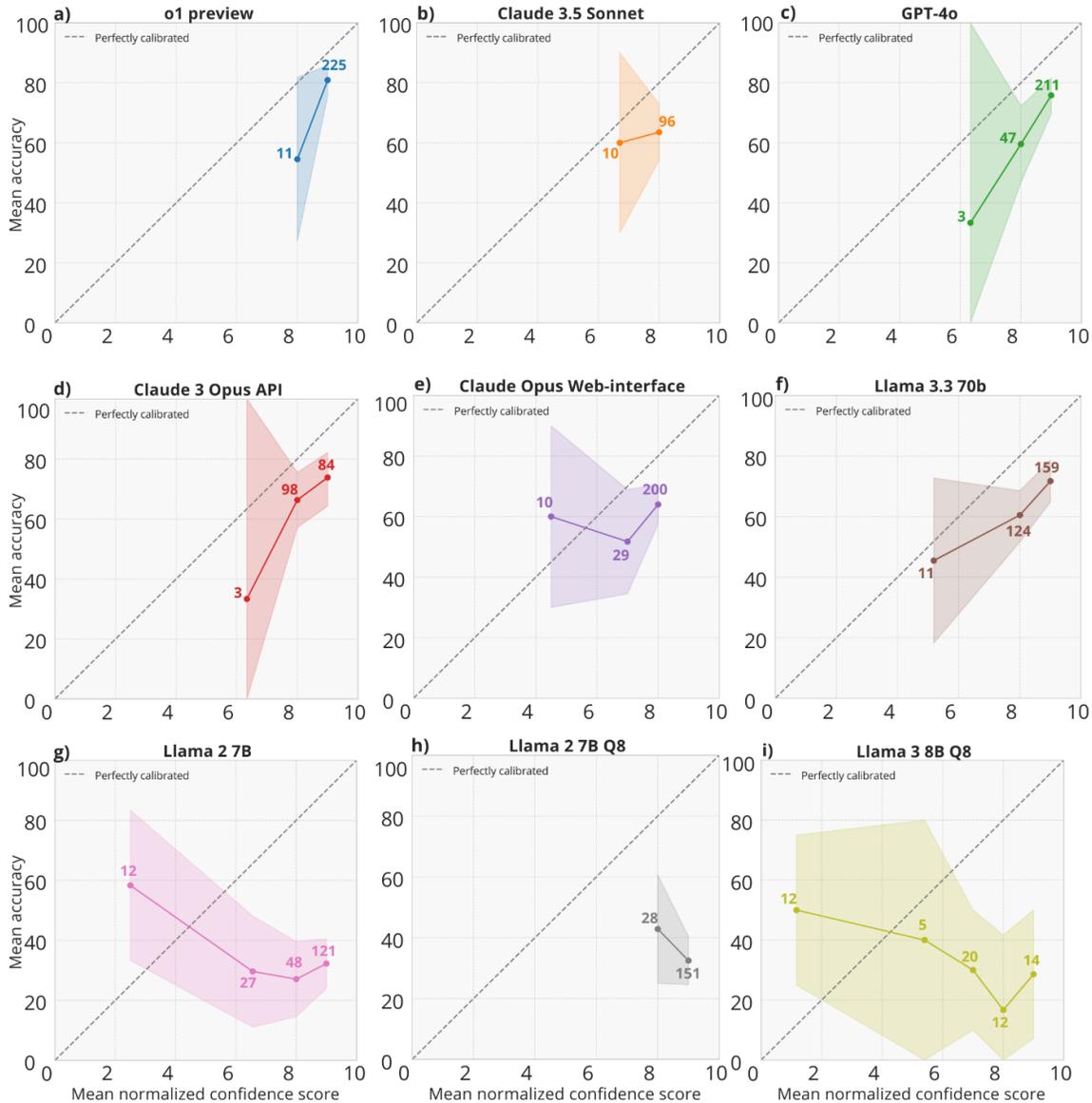

**Figure 4. Calibration curves for the top six (a-f) and bottom three (g-i) models.** Confidence scores were binned into intervals of 0.1 across the range of 0 to 1, with the mean normalized confidence score for each bin plotted against the corresponding observed accuracy. The dashed line represents perfect calibration.

We also conducted a stratified analysis of confidence scores by question difficulty and length, examining the three best (lowest Brier score) and three worst (highest Brier score) calibrated models. The LLM difficulty remained high for all question difficulty levels, but the accuracy decreased as the question difficulty increased (**Figure 5**). Confidence scores remained stable across all question lengths, and there was no noticeable change in response accuracy (**Supplementary Figure S3**).

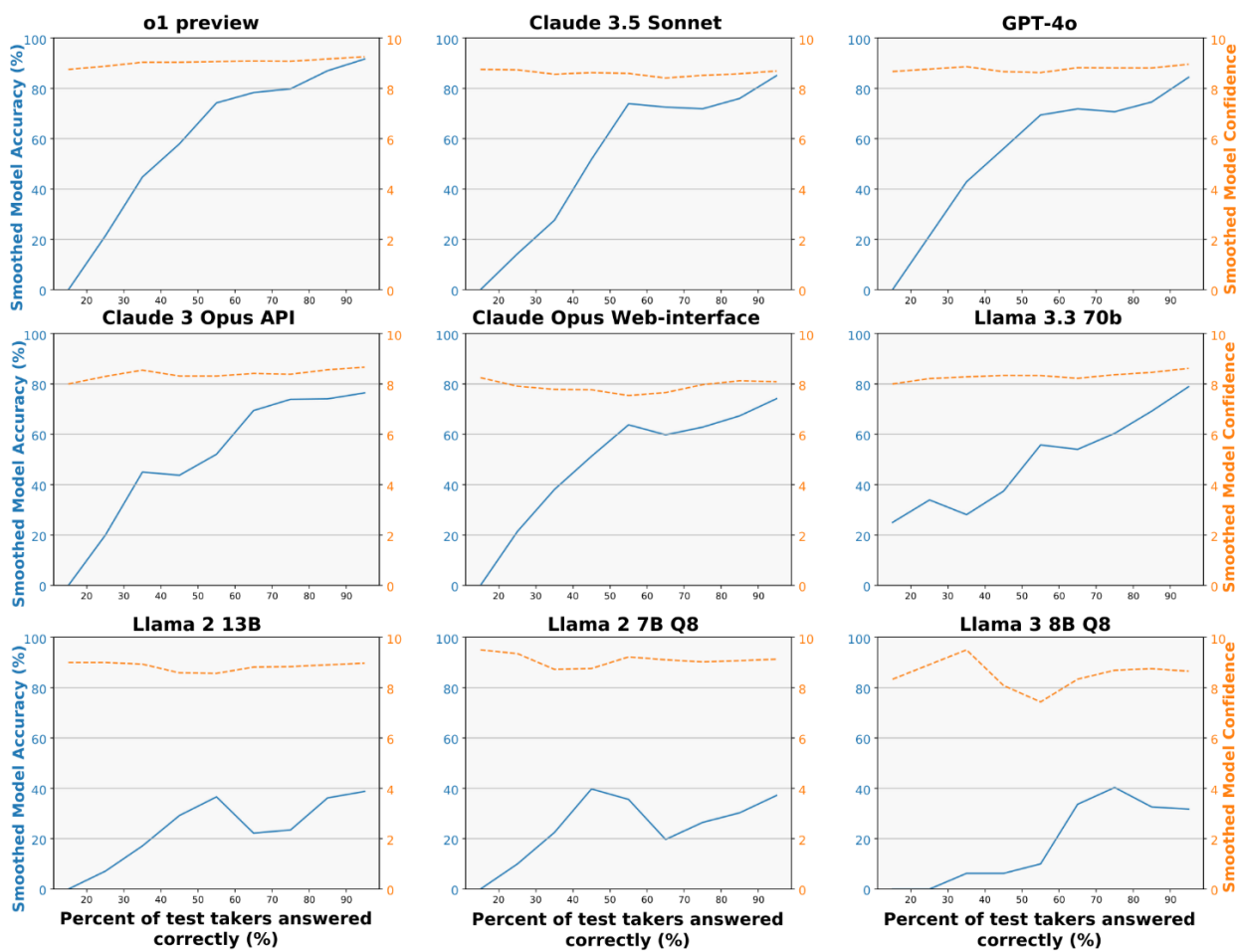

**Figure 5.** Subfigures (a) to (f) illustrate the smoothed trends in accuracy and confidence scores of large language models (LLMs) as a function of question difficulty, defined as the percentage of test-takers answering correctly. The questions were grouped into bins at 5% intervals to facilitate visualization. Across all models, confidence scores remained relatively stable despite increasing question difficulty and lower model response accuracy. Figures (a)–(c) highlight the three models

with the lowest Brier scores (highest calibration), whereas Figures (d)–(f) display the three models with the highest Brier scores (lowest calibration).

Our study represents a systematic investigation into the uncertainty quantification of LLM families in gastroenterology, focusing on the base performance of LLMs. However, more advanced methodological refinements may yield improved uncertainty estimation performance. By incorporating two-phase verification, utilizing BERT-based models, and implementing prompt engineering techniques, enhanced results can be achieved, although these remain insufficient to satisfy clinical requirements [13,17].

Previous attempts to assess LLM self-reported confidence have demonstrated overconfidence in LLMs in the non-medical and medical domains [22,23]. A previous benchmark of various LLMs on 2000 publicly available MCQs from five medical specialties also confirmed the lack of LLMs' ability to declare uncertainty. As the model accuracy in answering questions increases, its performance for verbal uncertainty quantification improves but remains unacceptable.

Although confidence elicitation underperforms alternative techniques, our study, along with Omar et al. 's[22], demonstrated that this capability improves with newer iterations and generations of models. This finding supports the hypothesis that by including datasets for this specific task in LLM training, these models can acquire certainty quantification skills. LLMs, particularly those of a larger scale, have the potential to acquire skills through training, a phenomenon known as emergence[24]. However, there is limited evidence regarding their capability for this specific skill, and further research is required.

It is also worth noting that the correlation between increased confidence and improved accuracy across model statistics suggests that better calibration scores may be the result of increasing accuracy, not decreasing accuracy, which is consistent with the findings of Yang et al[25].

Several limitations should be acknowledged in the interpretation of our results. Clinical MCQ, in particular, poses a distinct challenge for certainty estimation, as incorrect responses can still have varying degrees of correctness. Future frameworks must incorporate this fluid nature of medical ground truth, which is more complex than simple MCQs, when developing robust certainty estimation methodologies. Our prompt engineering strategy, while optimized for accuracy through expert emulation, may have inadvertently promoted model overconfidence. Another significant limitation of our study is the exclusive use of the GI dataset, which restricts the generalizability of our findings to the broader field of medicine. Additionally, while MCQ is an effective method for assessing the knowledge of both humans and LLMs, it does not fully capture the complexity of real-life scenarios, which are often more analogous to open-ended questions. Furthermore, although our dataset is restricted by a paywall, we cannot entirely rule out the possibility of data leakage or the incorporation of our dataset into the training process of LLMs.

In conclusion, our findings reveal persistent challenges in self-reported confidence in LLMs, characterized by limited discrimination capabilities and systematic overconfidence. Although successive generations of LLMs exhibit improved calibration, their performance remains insufficient for high-stakes medical applications. Targeted training protocols that explicitly address confidence estimation may enhance the utility of self-reported confidence. Ongoing research should focus on strategies such as few-shot injection and in-context learning to generate more reliable uncertainty estimates, ultimately improving model trust and explainability.

# Methods

## Reference dataset

The 2022 American College of Gastroenterology (ACG) self-assessment consists of 300 questions, of which 138 contain images. These questions were developed by a committee of gastroenterologists to reflect the knowledge, skills, and attitudes required for excellent patient care, covering a broad range of topics, including liver, colon, esophagus, pancreaticobiliary, and endoscopy. The questions were designed to assess higher-order thinking skills and were primarily case based. They were validated through statistical analysis of test-takers' performance, with an average correctness rate of 74.52% ± 19.49% on the 2022 assessment, indicating a moderate level of difficulty. Only the text portions of the questions and answers were used in this study's analyses. Questions were categorized by length (token count), difficulty (percentage of correct answers by test-takers), and patient care phase (treatment, diagnosis, or investigation). Additional details are provided in the Supplementary Section.

## Response Generation and Confidence Elicitation

We have previously described our methodology for question-response generation.[12] Briefly, 60 questions from the 2023 self-assessment exam and GPT-3.5 were used to select the model settings (temperature, maximum input, and output token count), prompt structure, and output format of all models. The configuration that maximized response accuracy was a temperature of 1, maximum token count of input token count + 512 output tokens, structured output approach, and prompt, as shown in figure (**Figure 1**). Among the various prompt engineering techniques evaluated, the following were identified as having a positive impact on the outcomes: expert mimicry, contextual embedding, Answer and Justify, Chain of Thought, confidence scoring, and direct questioning. OpenAI Web interface, OpenAI API, Claude Web interface, Claude API, Gemini Web interface, Poe Web interface, Firework API, and locally hosted hardware configurations such as RTX4090Ti and H100 systems were used for response generation and confidence elicitation.

## Confidence Score Extraction (Output Parsing)

To efficiently extract response and confidence data from the LLM outputs, we developed a structured output pipeline using GPT-4o (**Figure 1**). Our hybrid methodology combined regex-based rules to reduce the number of input tokens and LLM-based extraction to effectively parse the key portions of the LLM outputs. Briefly, sentences containing "confid*" are passed to an LLM-based extraction step that extracts either the certainty score (0-10) or classifies the score as "not_mentioned." Sentences classified as "not_mentioned" in the first pass are passed through the LLM-based extraction step a second time to maximize the extraction performance. The complete output parsing methodology is described in the Supplementary Material. To validate the output parsing pipeline, we compared it against manually extracted confidence scores from five randomly selected questions per model, achieving 98.8% accuracy.

Because some models did not reliably generate confidence scores, we excluded those that were missing confidence scores for more than 50% of the questions (Medicine-Chat Q8, OpenBioLLM-7B Q8, Qwen Qwq-32b, and GPT-3.5 Turbo). **Supplementary Figure S4** describes the distribution of missing confidence scores, with 30 models having missing confidence scores. **Supplementary Figure S5** illustrates a stratified analysis of response accuracy by confidence score missingness for models with missing scores for more than one-third of the questions.

## Statistical Analysis

We evaluated each model's performance from two perspectives: discrimination, the ability to distinguish between correct and incorrect responses, and calibration, the alignment between predicted confidence and actual accuracy.

Discrimination was quantified using AUROC. Specifically, we designated each response as 1 (positive) if it was labeled "correct" and 0 (negative) otherwise. The confidence scores of the model ranged from 0 to 10 and served as the continuous predictor variable. We employed the roc_auc_score function from sklearn.metrics to calculate the AUROC.

Conceptually, this involves varying the decision threshold over all possible confidence values, thereby classifying the responses as positive or negative at each threshold. For instance, let $t$ be a threshold such that:

$$Predicted\ class = \begin{cases} 1, & if\ confidence \geq t \\ 0, & if\ confidence < t \end{cases}$$

At each threshold, the true positive rate (TPR) and false positive rate (FPR) can be calculated as

$$TPR(t) = \frac{TP(t)}{TP(t) + FN(t)}$$

$$FPR(t) = \frac{FP(t)}{FP(t) + TN(t)}$$

where *TP, TN, FP, and F*N are the numbers of true positives, true negatives, false positives, and false negatives, respectively. By sweeping *t* across all confidence scores, we plotted TPR against FPR to form the ROC curve. The area under this curve yields a threshold-independent measure of how effectively the model discriminates between correct and incorrect responses, with higher AUROC values indicating stronger discriminative performance.

Calibration was evaluated using calibration plots, Brier score, and ECE. Calibration plots were generated by normalizing predicted confidence scores to a 0–1 scale, binning them into 0.1 intervals, and plotting the mean predicted confidence against the observed accuracy in each bin. Bins containing fewer than three predictions were excluded to ensure the reliability of the results. Bootstrap resampling (n = 1,000 iterations per bin) was used to derive 95% confidence intervals for each calibration point.

The ECE complements the Brier score by directly quantifying the aggregate discrepancy between predicted probabilities and observed outcomes across bins, whereas the Brier score measures the mean squared error between predictions and true labels. As a result, the Brier score reflects both calibration (how closely predicted probabilities match observed frequencies) and refinement (the sharpness of predictions), whereas ECE focuses more directly on calibration quality. Calculating both metrics provides a more comprehensive evaluation of model performance, capturing not only how well models are calibrated, but also the overall predictive accuracy of their probability estimates.

Our development and analysis were performed using Python 3.10. LLM answers were generated and extracted using the Openai Python library, Ollama application (v0.4), LM studio, and Langchain (v0.2 and v0.3). Statistical analyses were conducted using SciPy (v1.13.1) and Scikit-learn (v1.5.1), with data manipulation and visualization implemented through Pandas (v2.2.2), Matplotlib (v3.9.2), and Seaborn (v0.13.2).

# Ethical consideration

This study did not require ethical approval, as it did not involve human subjects or human data. We ensured data protection by confirming that the utilized LLM services did not retain or use our queries for model training purposes.

**Data availability**

The data supporting this study's findings were obtained from the American College of Gastroenterology (ACG) under license agreement. While these data are not publicly available owing to licensing restrictions, they may be obtained from the authors with the ACG's permission upon reasonable request. ACG self-assessment questions and answers are accessible to members through https://education.gi.org/.

**Code availability**

The underlying code for this study is available at https://github.com/narimannr2x/confidence_scoring.

**Acknowledgements**

This study was supported by the American Gastroenterological Association AGA-Amgen Fellowship-to-Faculty Transition Award (AGA2023-32-06) for AS. The funding source had no role in the study design, data collection, analysis, interpretation, or manuscript preparation.
We thank the American College of Gastroenterology for providing their question bank, the Hugging Face team for their accessible AI infrastructure, and the the Bloke account on Hugging Face for providing quantized versions of open-source LLMs. ChatGPT was used to assist with English language editing during manuscript preparation. The authors reviewed and edited all AI-assisted content and maintained full responsibility for the manuscript's content.

**Author contributions**

**Competing Interests**

The authors declare no competing financial interests or personal relationships that could have influenced the work reported in this study. NN: None; SAASN: None; AS: None; TS: None; GN: None; ZA: None; PL: None.

25. Yang, D., Tsai, Y.-H. H. & Yamada, M. On Verbalized Confidence Scores for LLMs. Preprint at https://doi.org/10.48550/arXiv.2412.14737 (2024).

# List of Supplementary Files

This is a supplementary file to **"Self-Reported Confidence of Large Language Model in Gastroenterology across Commercial, Open-Source, and Quantized Models"** by Nariman Naderi, Seyed Amir Ahmad Safavi-Naini, Thomas Savage, Zahra Atf, Peter Lewis, Girish Nadkarni, Ali Soroush.

Corresponding author: Ali Soroush (Ali.Soroush@mountsinai.org).

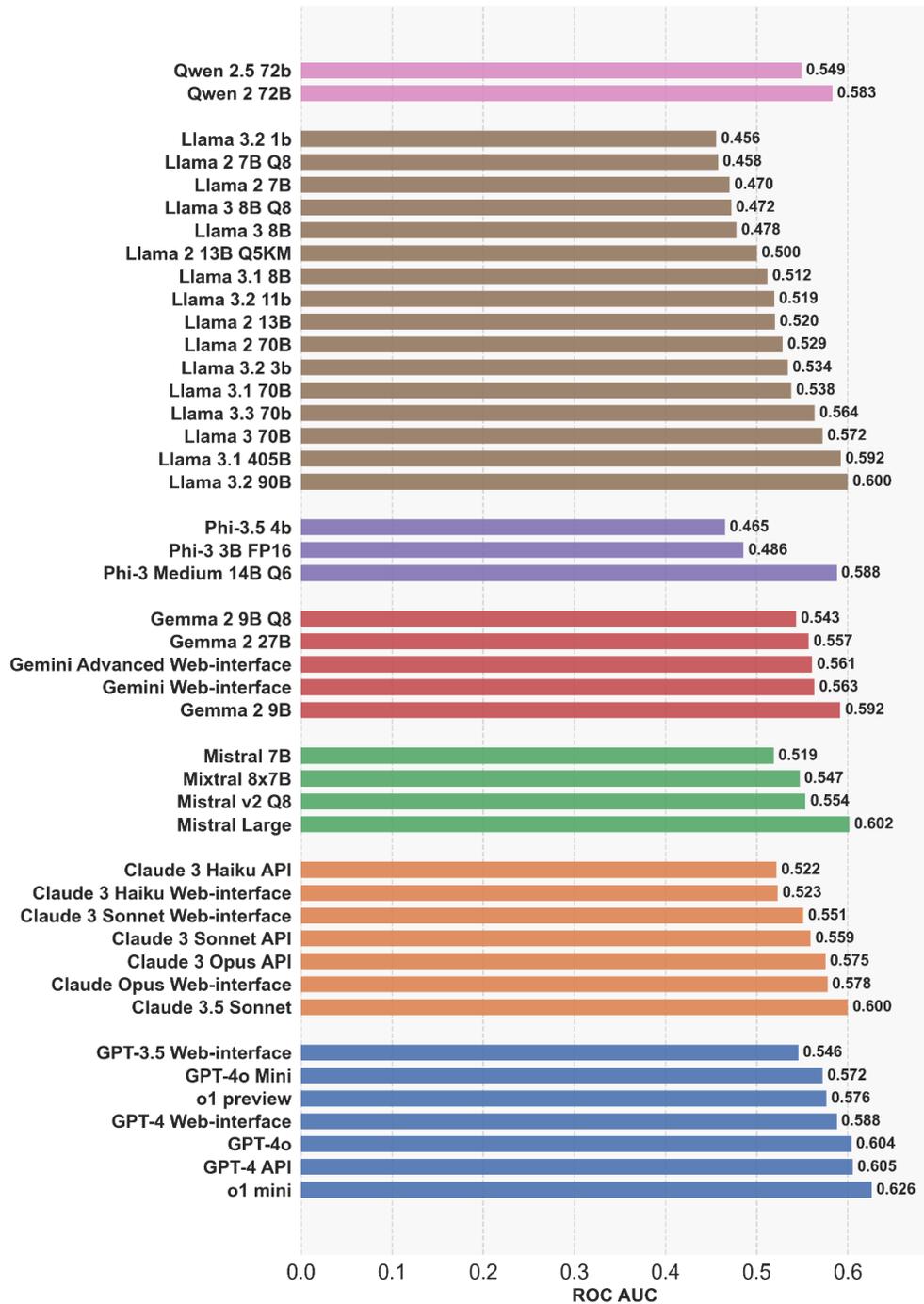

**Supplementary Figure S1.** Area Under the Curve of **Receiver Operating Characteristics (AUROC) Comparison of Various Model Families Based on Confidence Scores and Question Accuracy**. This graph presents the AUROC for each model, reflecting their performance in assigning confidence scores to questions relative to their accuracy. The models were grouped by their respective families for easier comparison. Receiver operating characteristic (ROC) curves were generated by comparing the model-derived confidence scores with binary correctness labels, and the area under the curve was computed to evaluate model discrimination.

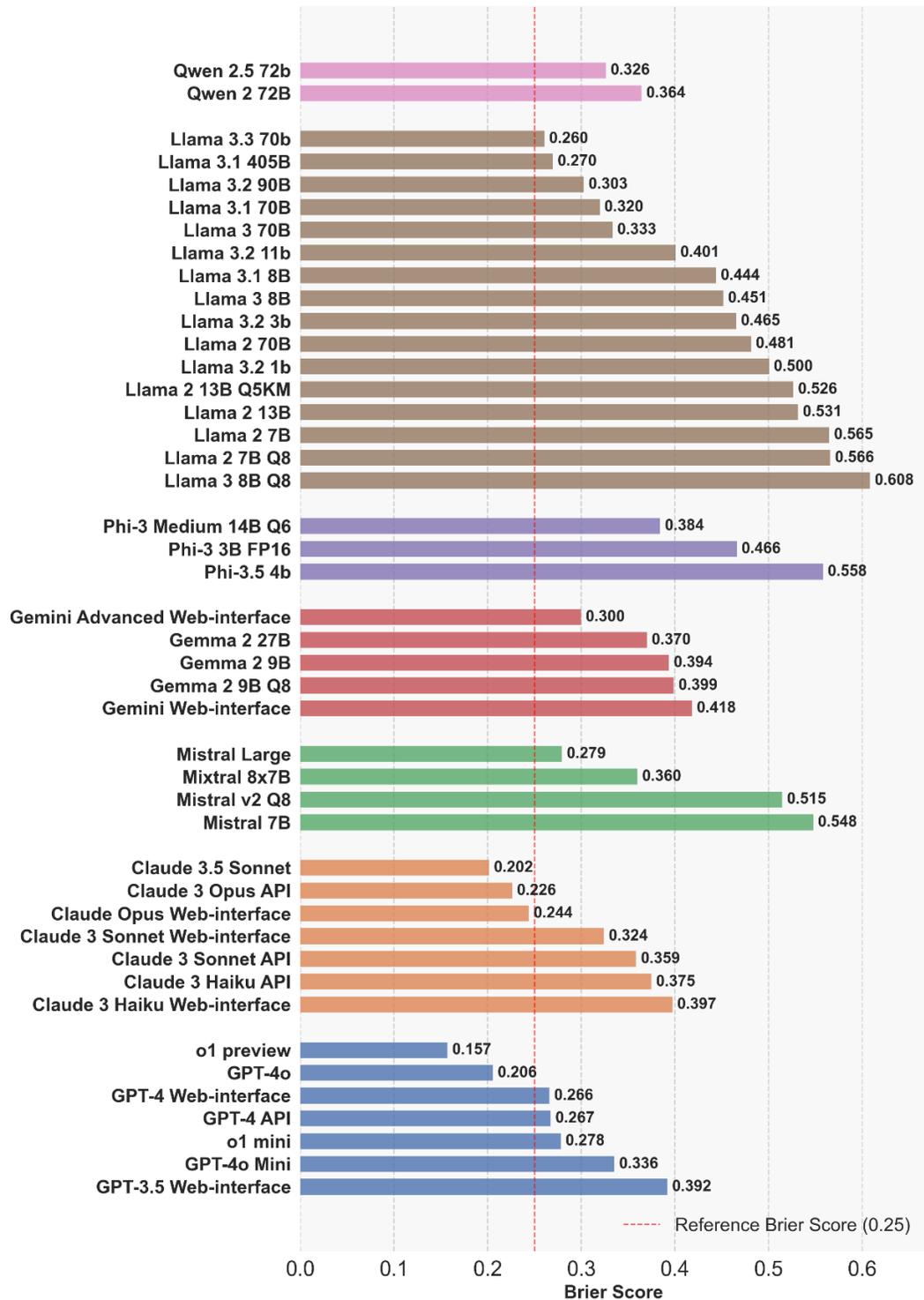

**Supplementary Figure S2. Brier Scores for LLM Confidence Elicitation.** The chart illustrates the comparative performance of the different language models, with lower scores indicating better calibration. The red dashed line indicates a reference Brier score of 0.25, representing the score expected from the random predictions. The models were grouped by their respective families for easier comparison.

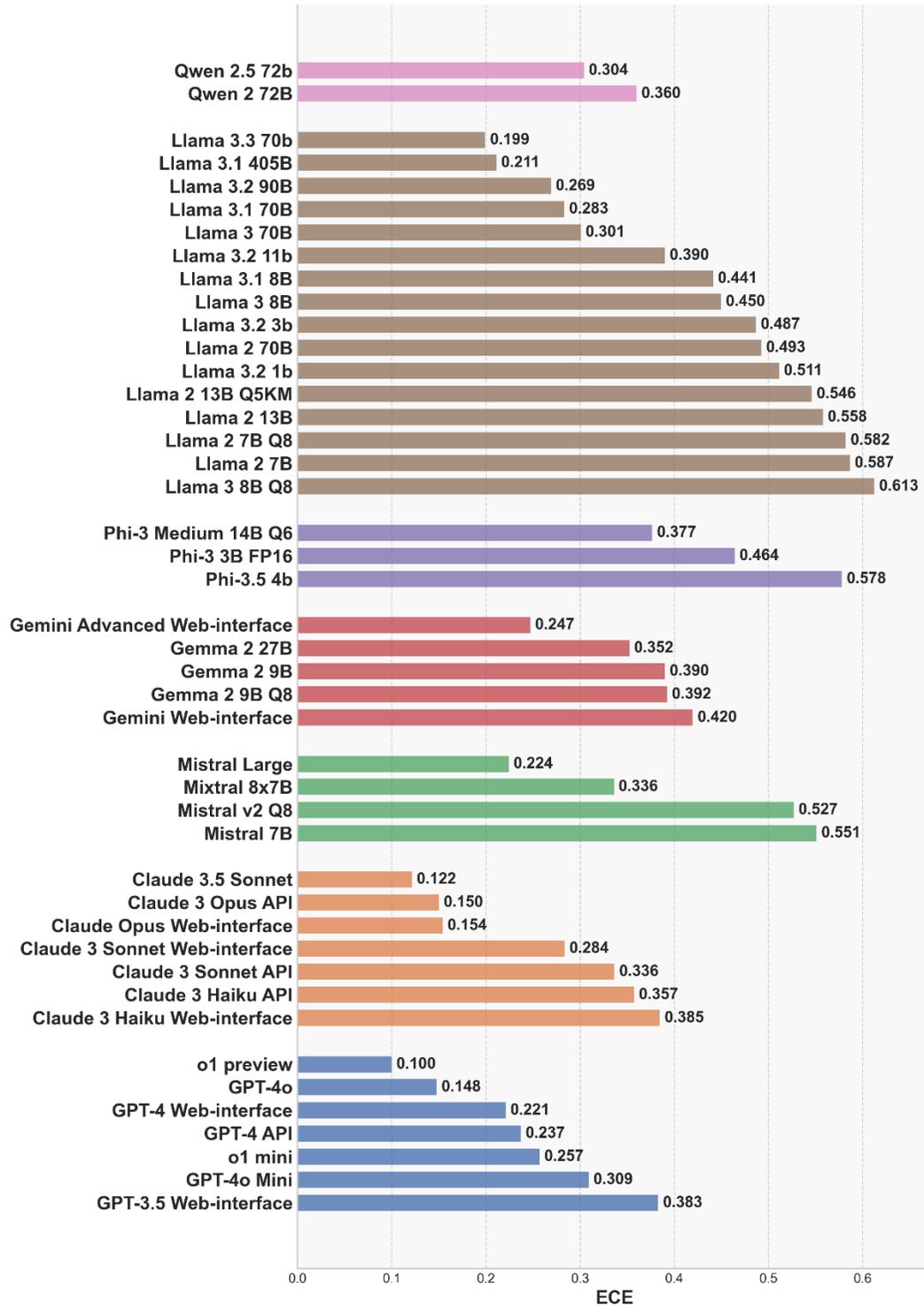

**Supplementary Figure S3. Expected Calibration Error (ECE) scores for LLMs in the context of confidence elicitation.** Lower scores indicate better calibration. Although there is no universally accepted threshold, an ECE value below 0.1 is commonly regarded as acceptable. Models are grouped by their respective families to facilitate comparison.

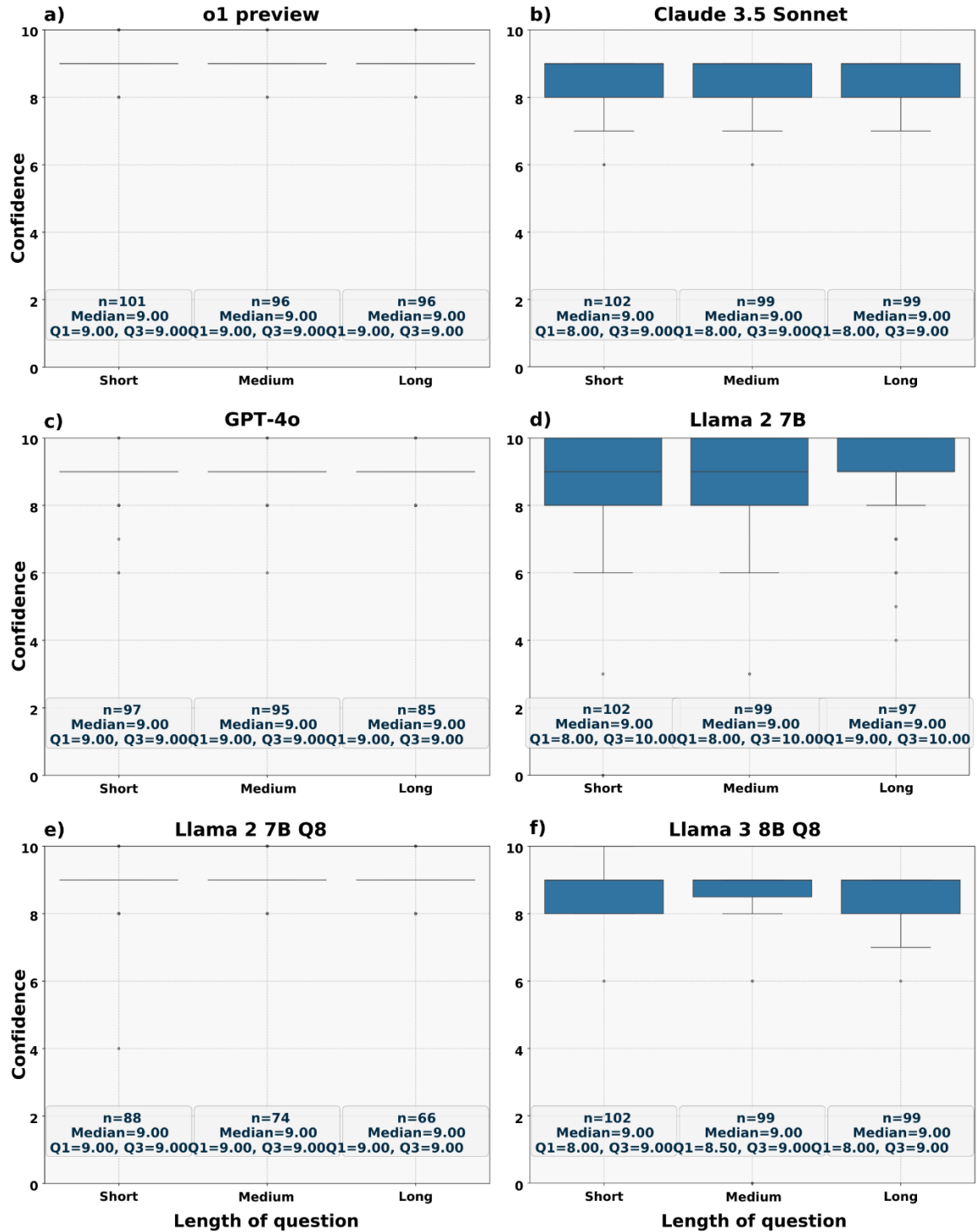

**Supplementary Figure S3.** Figures (a) to (f) present box plots illustrating the confidence scores elicited by the selected models stratified by question length. Response confidence scores appear qualitatively independent of the question length. Figures (a)–(c) highlight the three models with the lowest Brier scores (highest calibration), whereas Figures (d)–(f) display the three models with the highest Brier scores (lowest calibration).

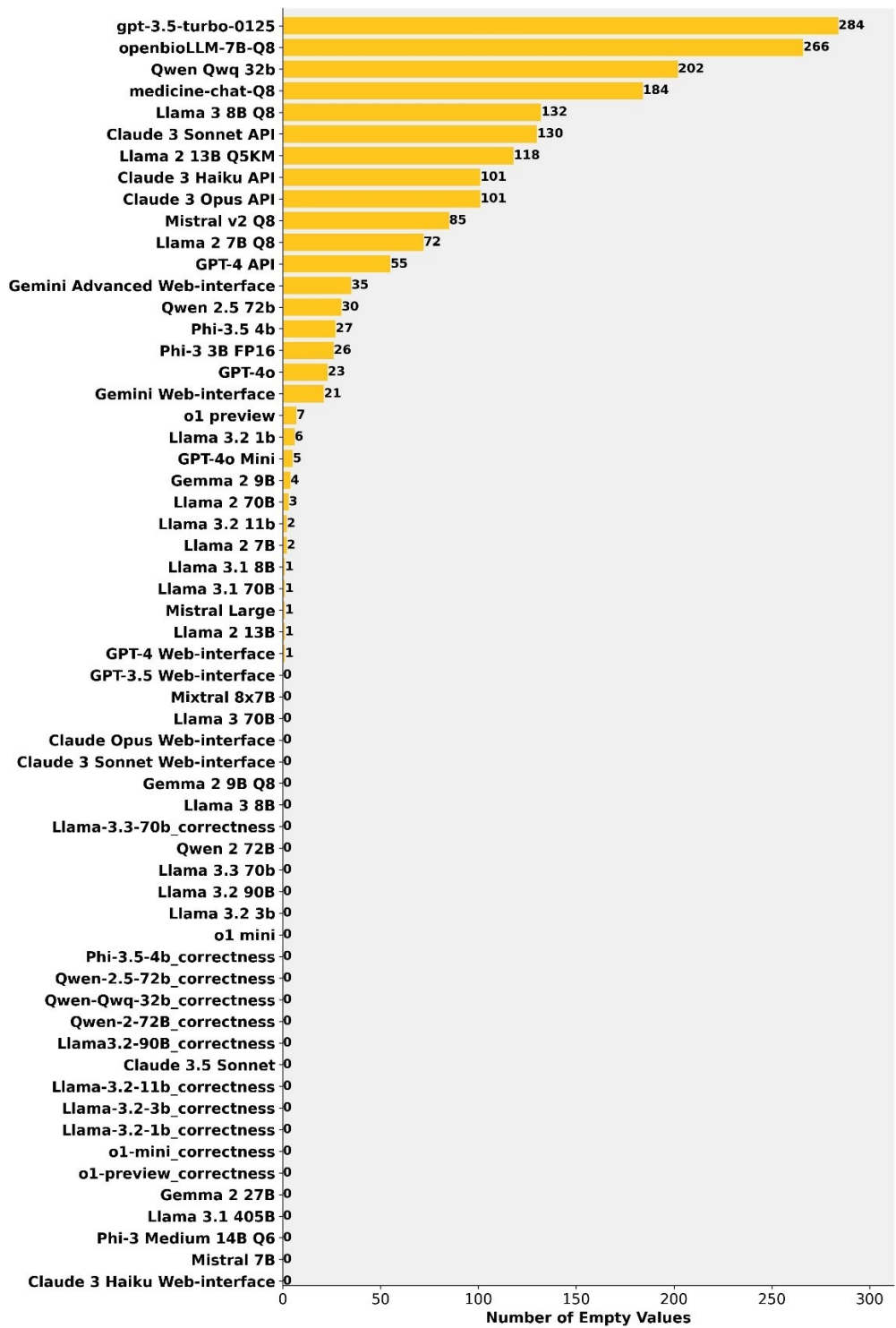

**Supplementary Figure S4. Non-generated confidence elicitation for each model, sorted from highest (top) to lowest (bottom).** Gpt-3.5-turbo-0125 exhibited the highest number of non-generations (n=284, 94.7%), followed by openbioLLM-7B-Q8 (n=266, 88.7%), and medicine-chat-Q8 (n=184, 61.3%).

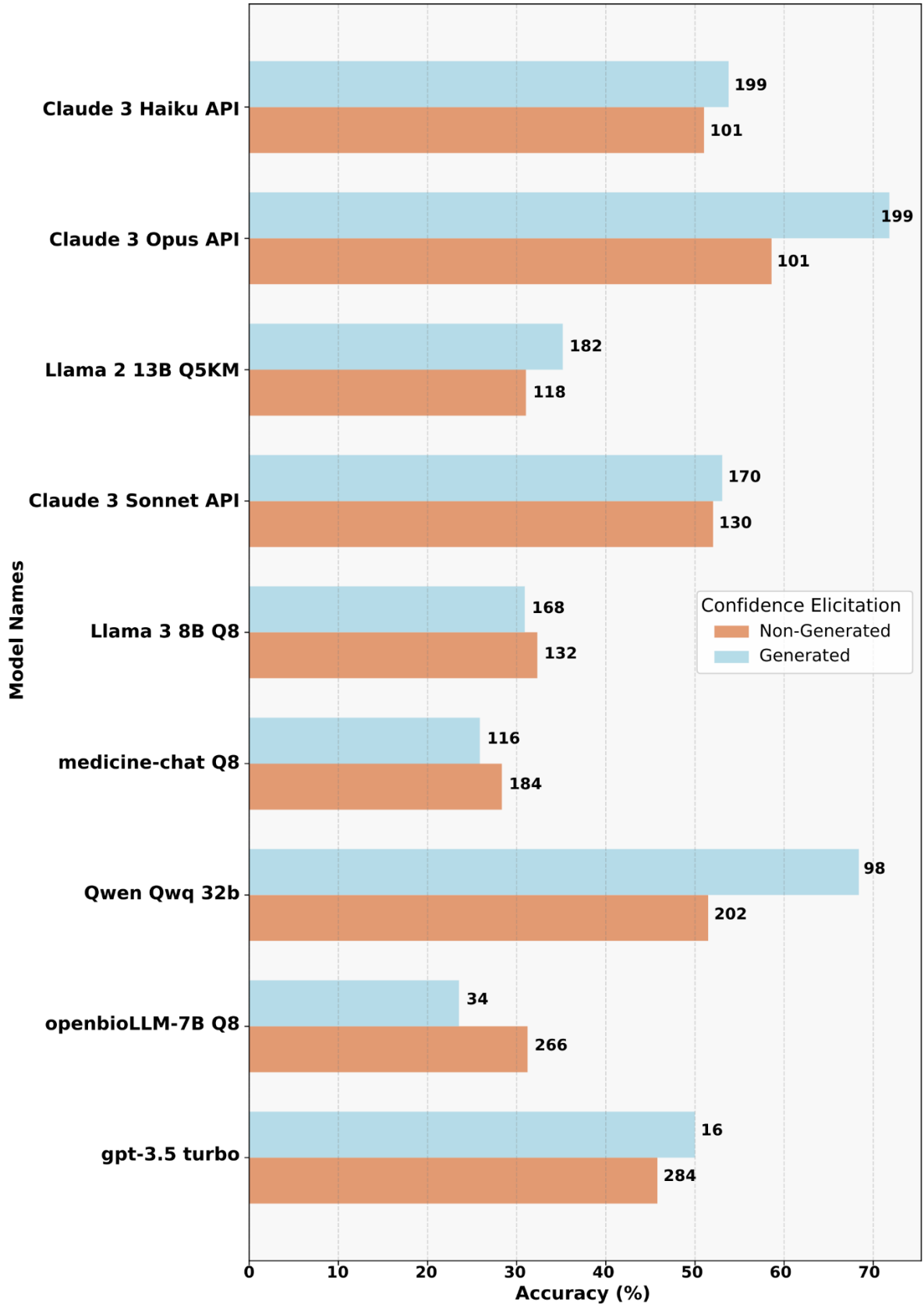

**Supplementary Figure S5. Model accuracy stratified by the generation of confidence elicitation**.

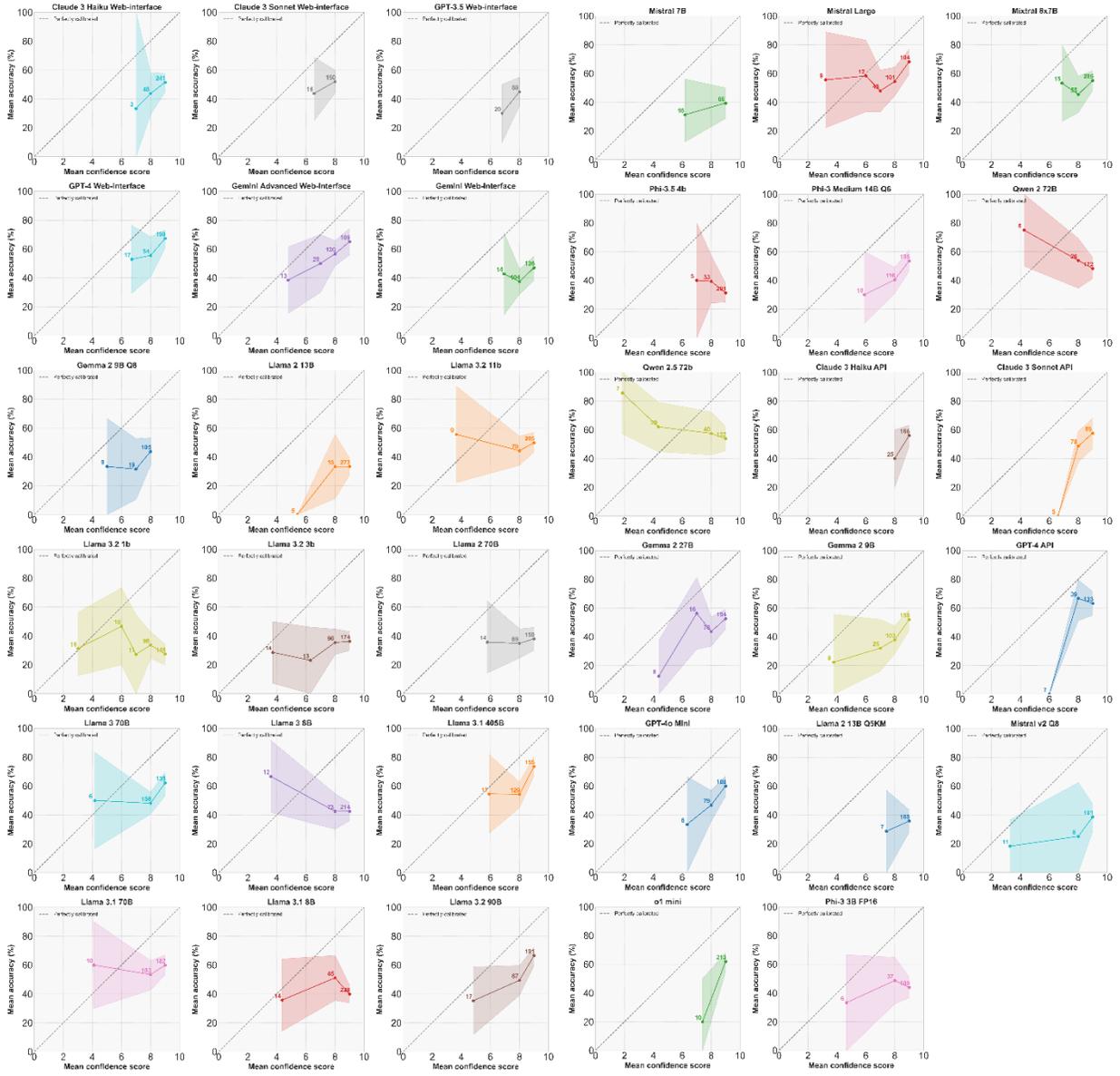

**Supplementary Figure S6. Calibration Curves for Middle-Tier Models.** This figure presents calibration curves for models with performance falling between the top six and bottom three models. Confidence scores were binned into intervals of 0.1 across the range of 0 to 1, with the mean normalized confidence score for each bin plotted against the corresponding observed accuracy. The dashed line represents perfect calibration.

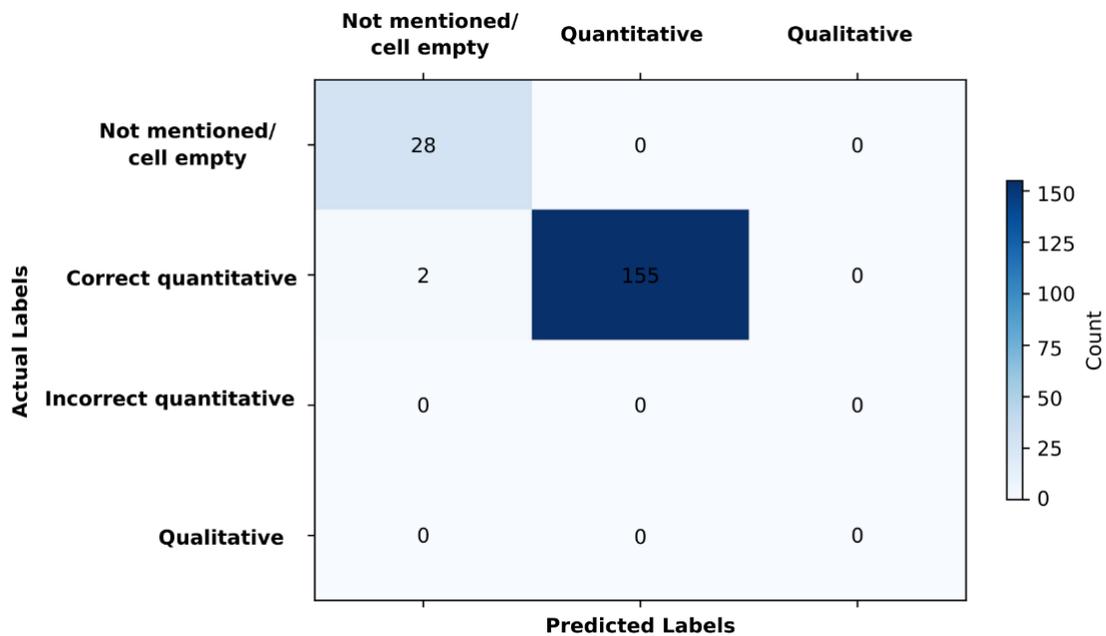

**Supplementary Figure S7. Confusion matrix of the accuracy of the automatic confidence extraction pipeline.** As mentioned in the text, we used an LLM extraction pipeline to extract the confidence numbers. Five questions were chosen from each model's answers for human evaluation. As stated above, the model accuracy was 98.91% (153 out of 155 questions).